\documentclass[prd,aps,twocolumn,a4paper,showkeys,nofootinbib,floatfix]{revtex4-1}

\usepackage{graphicx,psfrag}
\usepackage{mathrsfs}
\usepackage{amsmath,amsfonts,amssymb}
\usepackage{multirow}
\usepackage{comment}
\usepackage{tipa}
\usepackage{ulem}
\usepackage{color}
\usepackage{hyperref}
\usepackage{enumitem}
\usepackage{tipa}
\usepackage{bm}
\usepackage{algorithm}
\usepackage{algpseudocode}
\usepackage{lipsum}
\usepackage{aas_macros}

\newcommand{\be}{\begin{equation}}
\newcommand{\ee}{\end{equation}}
\newcommand{\bea}{\begin{eqnarray}}
\newcommand{\eea}{\end{eqnarray}}
\newcommand{\bel}{\begin{align}}
\newcommand{\eel}{\end{align}}

\hypersetup{
    colorlinks=true,
    linkcolor=red,
    citecolor=blue,
    urlcolor=black
}

\def\p{\partial}

\def\GMc2{{\rm G M_{\odot} c^{-2}}}

\def\L{\mathcal{L}}

\def\Z{{Z}}
\def\Zobs{\zeta}
\def\prf{\text{q}}
\def\hf{\mathfrak{h}}
\def\gf{\mathfrak{g}}

\def\params{\boldsymbol{\phi}}

\def\d{\text{d}}

\def\real{\boldsymbol{x}}
\def\latent{\boldsymbol{y}}
\def\pr{{\rm p}}
\def\normal{{\rm n}}
\def\f{\textbf{f}}

\def\Rosen{Rosenbrock }

\usepackage{pifont}

\usepackage{color}
\definecolor{cyan}{rgb}{0,0.6,0.6}
\definecolor{orange}{rgb}{0.9,0.5,0}
\definecolor{magenta}{rgb}{1,0,1}
\definecolor{purple}{rgb}{0.8,0.4,0.8}
\definecolor{gray}{rgb}{0.8242,0.8242,0.8242}

\definecolor{nicegreen}{rgb}{0.1,0.5,0.1}
\definecolor{nicered}{rgb}{0.7,0.1,0.1}

\begin{document}

\title{Bayesian evidence estimation from posterior samples with normalizing flows}

\author{Rahul \surname{Srinivasan}${}^{1,2}$}
\author{Marco \surname{Crisostomi}${}^{3,4}$}
\author{Roberto \surname{Trotta}${}^{1,2,5,6}$}
\author{Enrico \surname{Barausse}${}^{1,2}$}
\author{Matteo \surname{Breschi}${}^{1,2}$}
\affiliation{${}^{1}$SISSA, Via Bonomea 265, 34136 Trieste, Italy}
\affiliation{${}^{2}$INFN Sezione di Trieste, 34149 Trieste, Italy}
\affiliation{${}^{3}$IFPU - Institute for Fundamental Physics of the Universe, Via Beirut 2, 34014 Trieste, Italy}
\affiliation{${}^{4}$TAPIR, Division of Physics, Mathematics, and Astronomy, California Institute of Technology, Pasadena, California 91125, USA}
\affiliation{${}^{5}$Dipartimento di Fisica, Universit\`a di Pisa, Largo Bruno Pontecorvo 3, 56127 Pisa, Italy}
\affiliation{${}^{6}$Department of Physics, Imperial College London,  Blackett Lab, Prince Consort Rd,
SW7 2AZ London, United Kingdom}
\affiliation{${}^{7}$Italian Research Center on High Performance Computing, Big Data and Quantum Computing, 13 Via Magnanelli 2 – 40033, Italy}

\date{\today}

\begin{abstract}
We propose a novel method ({\it floZ}), based on normalizing flows, to estimate the Bayesian evidence (and its numerical uncertainty) from a pre-existing set of samples drawn from the unnormalized posterior distribution. We validate it on distributions whose evidence is known analytically, up to 15 parameter space dimensions, and compare with two state-of-the-art techniques for estimating the evidence: nested sampling (which computes the evidence as its main target) and a $k$-nearest-neighbors technique that produces evidence estimates from posterior samples. Provided representative samples from the target posterior are available, our method is more robust to posterior distributions with sharp features, especially in higher dimensions. For a simple multivariate Gaussian, we demonstrate its accuracy for up to 200 dimensions with $10^5$ posterior samples. 
{\it floZ} has wide applicability, e.g., to estimate evidence from variational inference, Markov Chain Monte Carlo samples, or any other method that delivers samples and their likelihood from the unnormalized posterior density. As a physical application, we use {\it floZ} to compute the Bayes factor for the presence of the first overtone in the ringdown signal of the gravitational wave data of GW150914, finding good agreement with nested sampling. 
\end{abstract}

\maketitle

\section{Introduction} 
\label{sec:intro}

One of the most important scientific tasks is that of discriminating between competing explanatory hypotheses for the data at hand. In classical statistics, this is accomplished by means of a hypothesis test, in which a null hypothesis (e.g., that the data do not contain a signal of interest) is rejected if, under the null, the sampling probability of obtaining a summary statistics as extreme or more extreme than what has been observed is small (the so-called $p$-value). An alternative --and often more fruitful-- viewpoint is offered by a Bayesian approach to probability, in which the focus is shifted away from rejecting a null hypothesis to comparing alternative explanations (see \cite{2008ConPh..49...71T} for a review).  This is done by means of the posterior odds (ratio of probabilities) between two (or several) competing models. The central quantity for this calculation is the Bayesian evidence, which gives the marginal likelihood for the data under each model once all the model parameters have been integrated out, tensioning each model's quality of fit against a quantitative notion of ``Occam's razor''. Bayesian model comparison has received great attention in cosmology (e.g.~\cite{Martin:2013nzq, Heavens:2018adv, 2017PhRvL.119j1301H}), and is being adopted as a standard approach to model selection in the gravitational wave (GW) community (e.g.~\cite{DelPozzo:2011pg, Vigeland:2013zwa, Toubiana:2021iuw, Roulet:2024cvl}) as well as in the exoplanets one (e.g.~\cite{Lavie_2017,2022ApJ...930..136L}).

The estimation of the Bayesian evidence is generally a challenging task, as it requires averaging the likelihood over the prior density in the model's parameter space. Various approaches have been proposed to this end. One that has gained particular prominence since its introduction by John Skilling \cite{Skilling2006} is nested sampling -- a method designed to transform the multi-dimensional integral of the Bayesian evidence into a one-dimensional integral. Since its invention, many different algorithmic implementations of nested sampling have been proposed, which in the main are characterized by the way the likelihood-constrained step is performed: these include ellipsoidal~\cite{2009MNRAS.398.1601F}, diffusive~\cite{2009arXiv0912.2380B}, dynamical \cite{2019S&C....29..891H} nested sampling; 
nested sampling with normalizing flows~\cite{Moss:2019fyi, Nessai_2021}; for a recent review, see~\cite{2023StSur..17..169B}). One of the remaining difficulties of nested sampling is its application to very large parameter spaces, where the curse of dimensionality hobbles most algorithms. Progress in this direction is being made thanks to approaches such as PolyChord~\cite{2015MNRAS.453.4384H} and proximal nested sampling \cite{2021arXiv210603646C}. 

Other methods to compute the evidence range from the analytical (Laplace approximation and higher-order moments~\cite{10.1214/16-EJS1218}, Savage-Dickey density ratio~\cite{Trotta:2005ar}) to the ones based on variations of density estimation, like parallel tempering MCMC coupled with thermodynamic integration~\cite{10.1063/1.1751356} and, more recently, on simulation-based inference  with neural density (or neural ratio) estimation and/or deep learning \cite{2023arXiv231115650K,Jeffrey:2023stk,2020arXiv200410629R}. 
Ref.~\cite{Mancini+2023} discussed evaluating the evidence using a normalizing flow and the ratio of a neural posterior estimator and neural likelihood estimator. However, the ratio of the two approximate quantities resulted in a larger compound error in the evidence. 

Another particularly interesting method was proposed in \cite{JiaSeljak_NFEvidence_2019}, where Optimal Bridge Sampling \cite{Meng1996, chen2012monte} is improved by using normalizing flows to obtain a suitable density estimator of the proposal distribution which has large overlap with the posterior.

A promising approach is that of evaluating the Bayesian evidence integral from a set of existing posterior samples, previously gathered e.g. via MCMC. The interest lies in the ability to obtain the evidence from the post-processing of posterior samples, which can be obtained with any suitable algorithm. Such a method, based on the distribution statistics of nearest neighbors of posterior samples, was presented in \cite{Heavens:2017afc} (see also Ref.~\cite{2021arXiv211112720M} for a similar approach based on harmonic reweighting of posterior samples, recently upgraded to a normalizing-flows based method in \cite{Polanska:2024arc, Piras2024TheFO}, and Ref.~\cite{Rinaldi:2024bow} for a hierarchical inference approach).
Here, we propose a new method based on normalizing flows.

Normalizing flows are a type of generative learning algorithms that aim to define a bijective transformation of a simple probability distribution 
into a more complex distribution by a sequence of invertible and differentiable mappings. 
Normalizing flows have been initially introduced in Ref.~\cite{Tabak:2010,Tabak:2013}
and then extended in various works,
with applications to clustering classification~\cite{Agnelli:2010},
density estimation~\cite{Rippel:2013,Laurence:2014},
and variational inference~\cite{Rezende:2015}.
While normalizing flows have been used in several parameter estimation scenarios in cosmology and GW physics \cite{Green:2020hst, Green:2020dnx, Dax:2021tsq, Dax:2021myb, 2023PhRvL.130q1403D, Wildberger:2022agw, Bhardwaj:2023xph, Crisostomi:2023tle, Leyde:2023iof, Shih:2023jme, Vallisneri:2024xfk}, here we introduce them, together with a novel training prescription, as a method to evaluate the evidence. This paper is structured as follows: we introduce normalizing flows in Section \ref{sec:method}, and how a suitable loss can be defined in order to encode the objective of evidence estimation; in Section~\ref{sec:results:valid_ndim}, we validate our approach on tractable likelihood in up to 15 parameter space dimensions, and benchmark it against {\tt dynesty} \cite{dynesty_code} (an implementation of nested sampling) and the $k$-nearest neighbors method of \cite{Heavens:2017afc}, demonstrating the superiority of our approach over the latter. In Section~\ref{sec:GW}, we apply our method to real GW data analysis by estimating the Bayes factor in favor of the presence of the first overtone in the ringdown phase of GW150914. We conclude in Section \ref{sec:conclusions} with an outlook on future applications.

\section{Method} 
\label{sec:method}

\subsection{Normalizing flows} 
\label{sec:method:flow}

Let $\real \in \mathbb{R}^d$ be a stochastic variable distributed according to 
a target probability -- in our case, the posterior -- $\real\sim\pr(\real)$, which can be arbitrarily complex.
Starting from a set of samples extracted from $\pr(\real)$,
normalizing flows allow one to map the 
variable $\real$ into a latent space, say $\latent\sim \normal(\latent)$,
where $ \normal(\latent)$ is an arbitrary tractable base distribution.
This transformation is performed training a neural network
in order to construct a bijection map $\f$ such that $\real =\f_{\params}(\latent)$,
where $\params$ are the network parameters that 
are defined by the form of the transformation
and require to be optimized.
Then, the target distribution can be mapped as
\be
\label{eq:flow}
\pr(\real) \mapsto \prf_{\params}(\real) =  \normal(\f_{\params}^{-1}(\real))\, \left|{\rm det}\frac{\p\f_{\params}^{-1}}{\p\real}(\real) \right|\,,
\ee
where $\f_{\params}^{-1}$ is the inverse transformation of $\f_{\params}$,
i.e. $\latent =\f_{\params}^{-1}(\real)$,
and ${\p \f_{\params}^{-1}}/{\p\real}$ is its Jacobian.
The transformation $\f$ can be arbitrarily complex, 
so that any distribution $\pr(\real)$ can be generated from any base distribution $\normal(\latent)$ 
under reasonable assumptions on the two distributions~\cite{Villani:2003,Bogachev:2005,Medvedev:2008}. 
In our implementation, we consider masked autoregressive flows (MAFs)~\cite{MAF}.

For our  purposes, we focus on the amortized variational inference approach
introduced in Ref.~\cite{Rezende:2015}, where 
the base distribution $\normal(\latent)$ is fixed to a simple probability function
(generally taken to be normal with zero mean and unit variance).
Within this approach, when the target is to learn the unnormalized posterior, the network parameters $\params$
are usually optimized with a loss function
defined as the cross entropy between the target $\pr(\real)$ and the reconstructed $\prf_{\params}(\real)$,
i.e.
\be
\label{eq:standard-loss} 
\begin{split}
\L_{1}(\params) &= -\int_{\mathbb{R}^d} \pr(\real) \, \log(\prf_{\params}(\real))\,\d\real\\
&=-{\rm E}_{\pr(\real)}\left[ \log(\prf_{\params}(\real))\right]\,,
\end{split}
\ee
where ${\rm E}_{\pr(\real)}$ represents the expectation value under $\pr(\real)$, which can be computed by a simple average given a set of independent samples.

\subsection{Evidence estimation} 
\label{sec:method:logz}

If the target distribution $\pr(\real)$ is defined up to an unknown normalization constant, $Z$, 
i.e. $\pr(\real)=\hat\pr(\real) / \Z$ and since $\pr(\real)$ is normalized by definition, we have,
\be
\label{eq:logz}
\Z=\int_{\mathbb{R}^d} \hat\pr(\real)\,\d\real\,.
\ee
In the context of Bayesian analysis, $\pr(\real)$ represents the posterior distribution of the parameters $\real$, $\hat\pr(\real)$ is the product of likelihood function and prior distribution (the unnormalized posterior), and $\Z$ is the evidence.

From Eq.~\eqref{eq:flow}, the evidence in Eq.~\eqref{eq:logz} can be written as 
\be
\label{eq:logz-flow} 
\Z\mapsto \Zobs(\real,\params) =  \frac{\hat \pr(\real)}{\prf_{\params}(\real)} \,,
\ee
from which it follows that
\be
\label{eq:logz-repar}
\Z \mapsto \frac{\hat\pr(\f_{\params}(\latent))}{\normal(\latent)} \, \left|{\rm det}\frac{\p\f_{\params}}{\p\latent} \right|=\frac{\hat\pr(\real)}{\normal(\f_{\params}^{-1}(\real))} \, \left|{\rm det}\frac{\p\f_{\params}^{-1}}{\p\real} \right|^{-1}\,.
\ee
This computation exploits the crucial property of $\normal(\latent)$ of being normalized by construction. In this paper, we evaluate $\hat\pr(\real)$ by multiplying the value of the likelihood function (saved in the output at each step of an MCMC chain) by the prior density at the same location. We note however that our method is generally applicable to any collection of posterior samples for which access to the unnormalized posterior $\hat\pr(\real)$ is available. For example, in the context of simulation-based inference, one could use neural conditional density estimation to learn the likelihood function \citep{doi:10.1073/pnas.1912789117}, then perform traditional MCMC with the neural likelihood to obtain samples from the posterior.

Given an unnormalized statistical model $\hat\pr(\real)$ 
and a transformation $\f_{\params}$,
the normalization $\Z$ can be explicitly computed as the 
right-hand side of Eq.~\eqref{eq:logz-flow}. 
Note that this computation can be performed in the variable $\real$
as well as in the latent space $\latent$,
as shown in Eq.~\eqref{eq:logz-repar}.

Moreover, while the {\it true} value of $\Z$ is constant,
the estimated $\Zobs$ is a function of the variable $\real$
and of the network parameters $\params$.
In practice, for a given set of $\params$ (after training),
the ratio in Eq.~\eqref{eq:logz-flow} will fluctuate according 
to the error of the trained flow $\f_{\params}$. We will address and model this uncertainty in the following Section.

\subsection{Additional loss terms}

Assuming we have an arbitrarily flexible bijection and a
combination of parameters $\params^*$ 
for which the transformation $\f_{\params^*}$ is exact, i.e. $\prf_{\params^*}(\real)=\pr(\real)$ $\forall \real$; 
then, the result of Eq.~\eqref{eq:logz-flow} is constant $\Zobs(\real,\params^*)=\Z$ $\forall\real$.
However, since the flux approximation is never exact, the estimated $\Zobs_i = \Zobs(\real_i,\params^*)$ at each of the posterior samples, $\real_i$, will be distributed around the true value of the evidence. 
While the minimum of Eq.~\eqref{eq:standard-loss}
is expected to satisfy this condition, we can additionally enforce this constraint as part of the loss to improve training.
This means that we are interested in the value of the network parameters $\params^*$
for which the distribution of $\Zobs$, $\hf(\Zobs|\params)$, results as close as possible to a Dirac delta function.
To this end, we define an additional loss function equal to the cross-entropy between the distribution $\hf(\Zobs|\params)$ 
(estimated with normalizing flows)\footnote{Notice that $\hf(\Zobs|\params)$ is not a probability distribution
in the proper sense; instead, it represents the histogram of the evidence obtained at each posterior sample value from the flow approximation, $\prf_{\params}(\real_i)$.}
and a $\delta$ function centered on the {\it true} value $\Z$, i.e.
\be
\label{eq:modified-loss}
\mathcal{L}_2(\params) = - \int \delta(\Zobs-\Z) \log\hf(\Zobs|\params) \d\Zobs  \,. 
\ee
Although 
$\Z$ is unknown and $\hf(\Zobs|\params)$ cannot be evaluated point-wise, we can still estimate expectation values for $\hf(\Zobs|\params)$
from the training samples.

Assuming that $\hf(\Zobs|\params)$ can be expanded in terms of its cumulants in a basis of Hermite polynomials, whose leading-order term is a normal distribution,
 from Eq.~(\ref{eq:modified-loss}) we obtain
\be
\mathcal{L}_2{(\params)}\simeq \log \sigma_\hf + \frac{(\mu_\hf - \Z)^2}{2\, \sigma^2_\hf} + {\cal O}(\mu_\hf-\Z)^3 \,, \label{eq:modified-loss-approx}
\ee
where $\mu_\hf$ and $\sigma^2_\hf$ are respectively the estimates of the first and second moment of $\hf(\Zobs|\params)$, derived from the training samples.
Therefore, provided that $|\mu_\hf - \Z| < \sigma_\hf$, namely that the {\it ground-truth} of $\Z$ is sufficiently close to the mean of the evidence distribution, we can neglect the quadratic term and approximate the loss function~\eqref{eq:modified-loss} with the standard deviation of the estimated evidence values across the samples:
$\mathcal{L}_2{(\params)}\simeq \log \sigma_\hf$. 

$\mathcal{L}_2$ helps to minimize the standard deviation of the evidence's distribution, but does not provide any information about the mean, as $\Z$ is unknown. To improve this feature, we consider
the distribution of the ratio estimated for pairs of samples 

\be
\rho = \frac{\Zobs(\textbf{\textit{x}}_i, \params)}{\Zobs(\textbf{\textit{x}}_j, \params)} \,,
\ee
which ideally would be a $\delta$ function centered at unity.
Thus, we can define a third loss function equal to the cross-entropy between the distribution of the evidence ratios $\gf(\rho|\params)$ and $\delta(\rho-1)$: 
\be
\label{eq:l3}
\mathcal{L}_{3}(\params) = - \int \delta(\rho-1) \log\gf(\rho|\params) \d\rho \,. 
\ee

Applying the same considerations that led to Eq.~(\ref{eq:modified-loss-approx}), we now obtain
\be
\mathcal{L}_3{(\params)}\simeq \log \sigma_\gf + \frac{(\mu_\gf - 1)^2}{2\, \sigma^2_\gf} + {\cal O}(\mu_\gf -1)^3 \,, \label{eq:modified-loss-approx3}
\ee
which can be broken into two separate losses

\bea
\mathcal{L}_{3a}(\params) &=& |\mu_\gf - 1|\approx |\log \mu_\gf| \,, \\[2ex] \label{eq:l3a}
\mathcal{L}_{3b}{(\params)} &=& \log\sigma_{\gf} \,. \label{eq:l3b}
\eea
where $\mu_\gf$ and $\sigma_\gf$ are estimated from the training samples via the sample average and standard deviation of $\rho$. Notice that, $|\mu_\gf - 1| < \sigma_\gf$ is required for the approximation (\ref{eq:modified-loss-approx3}) to hold.
A distinguishing feature of $\mathcal{L}_\mathrm{3}$ is that, using pairs of samples for the evaluation, their number is proportional to the square of the number of samples used to evaluate $\mathcal{L}_1$ and $\mathcal{L}_2$. This introduces statistical robustness to the training, especially for small training sets.

In our implementation, we exploit all the loss functions described above, following a scheme that we present in the next section (Sec.~\ref{sec:method:algorithm}). This provides higher accuracy than using any of the loss terms uniquely or in combination.

\subsection{Algorithm} 
\label{sec:method:algorithm}

Given a set of samples $\{\real_i\}$, for $i=1,\dots,N$,
extracted from $\pr(\real)$ and the corresponding unnormalized 
probability values $\{\hat\pr(\real_i)\}$, we proceed as follows:

\begin{figure*}
    \centering
    \includegraphics[width=.7\linewidth]{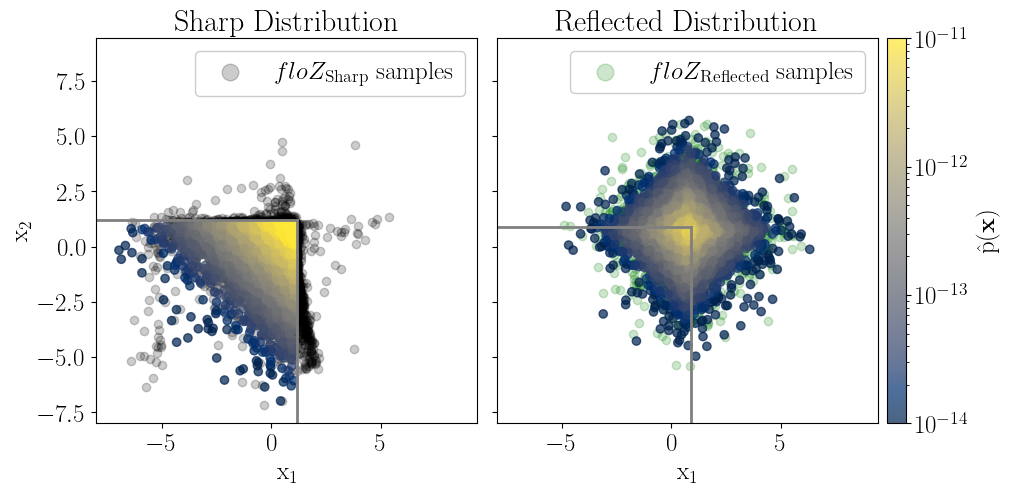}
    \caption{Effect of high target density near prior boundaries on the normalizing flow prediction. The color plot compares the original, ``sharp'' boundary distribution (left) with the reflected distribution (right). The color bar represents the un-normalized posterior probability density. The grey lines highlight the boundary. The background black (left) and green (right) scatter points are the predicted distribution by the flow trained on the respective distributions.}
    \label{fig:Reflection_Samples}
\end{figure*}

    \begin{figure}
        \centering
        \includegraphics[width=.8\linewidth]{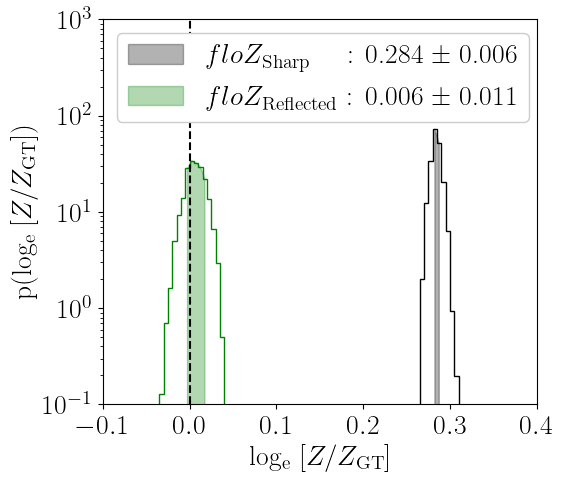}
        \caption{Effect of sharp boundaries on the evidence estimation. The black (green) probability density shows the distribution of the log evidence prediction from the original (reflected) samples. The log evidence is re-scaled by the ground truth, so the true value is at 0 (dashed black line).}
        \label{fig:Reflection_Evidence}
    \end{figure}

\begin{enumerate}
    \item {\it Pre-processing of input data:}
	We prepare the samples for training and validation. The two datasets contain independent samples. We whiten the samples to enhance the training process~\cite{Heavens:2017afc}. We do so by projecting the samples along the eigenvectors of the covariance matrix of the samples and rescaling by the square root of the eigenvalues to ensure unit variance. We assign 80\% of the generated whitened MCMC samples to the training dataset, and the rest to the validation dataset.
 
    \item {\it Training the neural network:}
        We train the normalizing flow on the training set employing gradient-based optimization~\cite{Kingma:2014}. 
        The loss function cycles through the four previously discussed loss functions ($\mathcal{L}_1$, $\mathcal{L}_2$, $\mathcal{L}_{3a}$, $\mathcal{L}_{3b}$). As done in the framework of transfer learning, the network is trained on one loss before being adapted to minimize the next. Initially, with $\mathcal{L}_{1}$, the model is trained to learn the general features of the posterior distribution to provide a first estimate of the evidence -- which is necessary to satisfy the assumptions on which $\mathcal{L}_{2}$ relies. Then, $\mathcal{L}_{2}$ incentivizes the pre-trained model to reduce the error in the evidence estimation. $\mathcal{L}_{3a}$ and $\mathcal{L}_{3b}$ sequentially reduce the disparity in the network's evidence prediction from each sample. This cyclic training regime, wherein each loss function is applied sequentially, repeats every $N_e$ epoch. We schedule the losses  as follows:

    \be
    \label{eq:loss}
    \L(\params) \equiv \begin{cases} 
        \L_1(\params) & e \in [0, 0.25-t_e), \\
        \alpha \L_1(\params) + (1-\alpha) \L_2(\params) & e \in [0.25-t_e, 0.25), \\
        \L_2(\params) & e \in [0.25,0.5-t_e), \\
        \alpha \L_2(\params) + (1-\alpha) \L_{3a}(\params) & e \in [0.5-t_e,0.5), \\
        \L_{3a}(\params) & e \in [0.5,0.75-t_e), \\
        \alpha \L_{3a}(\params) + (1-\alpha) \L_{3b}(\params) & e \in [0.75-t_e,0.75), \\
        \L_{3b}(\params) & e \in [0.75,1-t_e), \\
        \alpha \L_{3b}(\params) + (1-\alpha) \L_1(\params) & e \in [1-t_e,1), \end{cases}
    \ee
    
    where $e := (\mathrm{epoch} \bmod N_e)/N_e$ represents the fractional training epoch, $t_e$ is the fractional transition period ($0 < t_e < 0.25$) that determines the smoothness of the transition between one loss term to the next, and $\alpha := \min(1, \max(0, (0.25 - e \bmod 0.25)/t_e))$.  
    
    Each loss term uniquely contributes to $0.25 - t_e$ fractional epochs before transitioning to the next term. The transition is defined by the continuous linear equation of the form $\alpha \L_\mathrm{prev} + (1-\alpha) \L_\mathrm{next}$, where $\L_\mathrm{next}$, $\L_\mathrm{prev}$ are respectively the next and previous loss terms, and $\alpha$ ranges from 1 to 0 with increasing epochs (as specified above). A sudden transition ($t_e \ll 0.25$) results in an oscillatory behavior, wherein the loss fluctuates between the local minima of each loss term without net change. However, a slow transition ($t_e \sim 0.25$) results in slow training, due to reduced training time allocated for each loss term separately. Empirically, we find the optimal training rate for a training cycle period $N_e = 100$, with a transition period $t_e = 0.05$ (corresponding to 5 epochs).

    As an alternative to the adaptive loss weighting, we tested another formulation of the loss function, wherein the weights of the weighted sum of the individual loss terms are optimized alongside network parameters $\params$:
    \be
    \label{eq:loss}
    \begin{split}
        \L(\params,\eta, \beta, \gamma, \delta) = \,&\eta \L_1(\params) + \beta \L_2(\params) \\
        &+ \gamma \L_{3a}(\params)+ \delta \L_{3b}(\params)\,,
    \end{split}
    \ee
    where $\eta, \beta, \gamma, \delta$ are optimizer-tuned weights with values that range from 0 to 1. However, the loss scheduler was often more accurate and sometimes faster to reach convergence by way of exceeding the \textit{patience}, especially for higher dimensions ($> 5$). 
    
    The training terminates when at least one of the following three conditions is satisfied:
	\begin{itemize}
		\item {\it Maximum iteration}: The training stops if the number of epochs exceeds a fixed number of maximum iterations. We fix this number to 500 for all cases studied in this paper. However, more complex distributions or higher dimensions could need a larger value. \\
        
		\item {\it Patience}: If the optimizer does not find improvement in the loss of the validation dataset after a fixed number of iterations, the training terminates. This is to prevent overtraining. We fix this number to 200.\\
		\item {\it Tolerance}: As in the case of nested sampling, it is possible to require that the algorithm achieves a predetermined accuracy (tolerance) in the evidence estimation. When the error on the estimated evidence is less than this threshold, the training is stopped.
        Given the exploratory nature of this paper, we never used this condition.
	\end{itemize}
 
	\item {\it Extracting evidence information:}
	Given the optimized set of network parameters $\params^*$, we estimate $\Zobs(\real_i,\params^*)$ according to Eq.~\eqref{eq:logz-flow}, over a subset of  training samples $\{\real_i\}$. Since $q_{\params^*}(\real_i)$ is more accurate in the bulk of the distribution, following Ref.~\cite{ImportanceSampling_DiCiccio+97}, we select the samples in the latent space $\{\latent_i\}$ within a sphere $\bm{\mathcal{B}}$ centered about zero and with Mahalanobis distance $\delta$, which in the case of the Gaussian latent space simplifies as $\bm{\mathcal{B}} = \{\latent_{\bm{\mathcal{B}}} \in \latent : ||\latent|| < \delta\}$. We choose $\delta=\sqrt{d}$, i.e. containing the 1-$\sigma$ region of the latent distribution.

\end{enumerate}

A perfectly trained flow transforms the real samples into a normal distribution with zero mean and unit variance in latent space, which in future work we will exploit to define a stopping criterion for training. This could be achieved by verifying that the flow's latent distribution is congruent to a standard Normal, for example, via the Kullback–Leibler divergence.

\subsection{Addressing limitations of normalizing flows}
\label{sec:methods:flowDrawbacks}

Although normalizing flows are very expressive in modeling complex distributions, in the context of evidence estimation we find they can be inefficient in two aspects.
\begin{itemize}
    \item \textit{Modeling distributions that lie far from the origin:} As the latent space is centered at the origin, the flow can require many training cycles to correct for the translation required to model the target real distribution. We reduce the training time by subtracting the mean value from the samples, thus re-centering the bulk of the real distribution to the origin.

    \item \textit{Modeling distributions with high density in proximity to boundaries:} Flows model a target distribution through a diffeomorphic transformation to the latent space. Therefore, they inherently struggle to model discontinuities, i.e. distributions with a high density near a sharp prior boundary. We circumvent this limitation by sequentially transferring half of the samples to their reflected position about each of the sharp edges, ultimately resulting in a continuous distribution better suited for the flow to model. About each sharp edge, the prior volume is doubled, thus halving the prior density. To account for the change, we rescale the un-normalized probability of all samples by a factor of 2 for each sharp edge.
    
    Figure~\ref{fig:Reflection_Samples}, left, shows one such distribution with peak density adjacent to two boundaries, marked by the grey lines. We perform successive sample transfers about both boundaries, resulting in the distribution on the right. Note that the un-normalized probability of the final distribution is one-fourth that of the original due to the quadrupling of the prior volume. The discrepancy between the target distribution (color plot) and the flow's predicted distribution (scatter points) is apparent in the original distribution, as illustrated by the leakage of the flow's predicted samples outside the prior boundaries. In contrast, the reflected distribution has flow samples consistent with the target, with no significant outliers.

    Figure~\ref{fig:Reflection_Evidence} compares the evidence estimation from the two distributions. The prediction from the original distribution significantly overestimates the evidence, while the estimation using reflected samples is well-centered on the ground truth.

    Our method of translating samples about sharp edges can be adapted for other techniques that estimate the evidence from samples, such as techniques that use nested sampling and k-nearest neighbor. We discuss this with an example in Sec~\ref{sec:results:benchmark} and Figure~\ref{fig:Comparison}.

   Finally, in the case of periodic parameters (such as angles), a distribution may have an over-density at the upper and lower boundary. We smoothly merge these densities by appropriately translating the parameter samples, modulo the periodicity.
    
\end{itemize}

\section{Validation and scalability}
\label{sec:results:valid_ndim}
In Secs.~\ref{sec:results:TractableEvidence} and~\ref{sec:results:IntractableEvidence}, we illustrate examples of posterior samples with tractable (i.e.~analytic) and intractable (i.e.~unknown analytically) evidence, respectively. Table~\ref{table:likelihoods} summarizes the different likelihood distributions used to obtain the posterior samples. Using these examples, in Sec.~\ref{sec:results:benchmark} we demonstrate the validity of our method and explore its dimensional scalability by benchmarking the results against existing evidence estimation techniques.

\subsection{Posterior samples with tractable evidence}
\label{sec:results:TractableEvidence}
To validate our technique, we design the following unnormalized posteriors, $\hat{p}(\real)$, with analytically tractable evidence. For simplicity, we choose a flat, rectangular prior that defines the finite boundaries of the unnormalized distribution.

\begin{itemize}
    \item {\it Truncated $d$-dimensional single Gaussian:}
    We define a multivariate unnormalized Gaussian posterior in $d$ dimensions, from which we draw samples truncated by the rectangular uniform prior. The distribution is defined in terms of a $d$-dimensional mean and a $d \times d$ covariance matrix. Given the multivariate Gaussian distribution of the latent variables, this posterior is simple for the normalizing flow to model, i.e. the normalizing flow obtained by minimizing the loss function is simply an affine transformation.
    
    \item {\it Truncated $d$-dimensional mixture of five Gaussian distributions:}
    The unnormalized posterior is a mixture model of five different multivariate Gaussians, with equal mixture weights. Each of the five Gaussians has a different mean and covariance matrix, and the distribution is truncated by the rectangular prior. The Gaussian mixture is a non-trivial extension of the single Gaussian case.

    \item {\it Truncated $d$-dimensional exponential function:}
    We define a multivariate, unnormalized posterior that follows an exponential shape along positive values of each parameter, and enforce a rectangular prior with one vertex at the origin. We assign random (but fixed) values to the components of the exponent $\bm{\lambda}$. The unnormalized posterior is peaked at the origin. The resulting sharp distribution is non-trivial to model, as discussed in Sec.~\ref{sec:methods:flowDrawbacks}. We further highlight the importance of addressing sharp posterior boundaries in Sec.~\ref{sec:results:benchmark}.
    
\end{itemize}

\subsection{Posterior samples with intractable evidence}
\label{sec:results:IntractableEvidence}
We also test the method with an unnormalized probability distribution with analytically intractable evidence for dimensions higher than 2. For dimensions higher than 3 the evidence is very expensive to compute even numerically, therefore we do not provide a ground truth as a reference.

\begin{itemize}
\item {\it{Truncated $d$-dimensional Rosenbrock distribution:}}
    The Rosenbrock function \cite{Rosenbrock1960AnAM} and its higher dimensional extensions described in Ref.~\cite{Goodman+2010} are often used to test the efficacy of MCMC  sampling algorithms \cite{Pagani+2021}, as it is generally hard to probe the maxima of the distribution. Moreover, the normalization constant is typically unknown. We use a $d$-dimensional Rosenbrock function as defined in Ref.~\cite{Goodman+2010} and shown in Table~\ref{table:likelihoods} to describe the unnormalized posterior.
\end{itemize}

\begin{table*}
\centering
\begin{tabular}{||l | c | c ||} 
 \hline
 Functional form & $\hat\pr(\real)$ & $Z$ \\ [.5ex] 
 \hline\hline
 & & \\
 Gaussian & $\exp\left[-\frac{1}{2}(\bm{\mu}-\bm{x})^T \bm{\sigma}^{-1}(\bm{\mu}-\bm{x})\right] I({\bm{x}}-{\bm{x}}_i) I({\bm{x}}_f- {\bm{x}})$ & $(2\pi)^{d/2}|\bm{\sigma}|^{1/2}\left[\mathrm{erf}(\bm{x}_i-\bm{\mu}, \bm{\sigma}) - \mathrm{erf}(\bm{\mu}-{\bf{x}}_f, \bm{\sigma})\right]$ \\[3ex]
Gaussian Mixture & $\sum_{j=1}^5 \exp\left[-\frac{1}{2}(\bm{\mu}_j-\bm{x})^T \bm{\sigma}_j^{-1}(\bm{\mu}_j-\bm{x})\right] $ & $\frac{1}{5}\sum_{j=1}^5 (2\pi)^{d/2}|\bm{\sigma}_j|^{1/2}$ \\
& $ I({\bm{x}}-{\bm{x}}_i) I({\bm{x}}_f- {\bm{x}})$ & $ \left[\mathrm{erf}(\bm{x}_i-\bm{\mu}_j, \bm{\sigma}_j) - \mathrm{erf}(\bm{\mu}_j-{\bf{x}}_f, \bm{\sigma}_j)\right]$ \\[3ex]
Exponential & $\mathrm{exp}\left[-\bm{\lambda} \, \bm{x}\right]  I({\bm{x}}) I({\bm{x}}_f- {\bm{x}})$ & $|\prod_{i=1}^{n} \lambda_i|^{-1}\left[1 - \mathrm{exp}(-\bm{\lambda} \, \bm{x}_f)\right]$ \\[3ex]
 \Rosen & $\mathrm{exp}\{-\Sigma_{j=1}^{d-1} [A(x_{j+1} - x_{j}^2)^2 + (1-x_j)^2]/B \}$ & \\ & $ I({\bm{x}}-{\bm{x}}_i) I({\bm{x}}_f- {\bm{x}})$ &\\ [3ex] 
 \hline
\end{tabular}
\caption{Summary of the unnormalized posteriors (and their integrals) used to validate and benchmark our evidence estimation. The lower and upper bounds of the rectangular prior are given by $\bm{{x}}_i, \bm{x}_f$, respectively, the indicator function $I(\bm{z}) = 1$ for $\bm{z} > \bm{0}$ and 0 elsewhere and $\text{erf}({\bm{x}}, \bm{\sigma})$ is the multivariate Normal error function. The vectors $\bm{\mu}, \bm{\mu}_j \, (j=1,\dots,5)$, $\bm{\lambda}$,
covariance matrices $\bm{\sigma}, \bm{\sigma}_j \, (j=1,\dots,5) \in \mathbb{R}^{d\times d}$, and real scalars $A,B$ are fixed.}
\label{table:likelihoods}
\end{table*}

\begin{figure*}
    \centering
    \includegraphics[width=.6\linewidth]{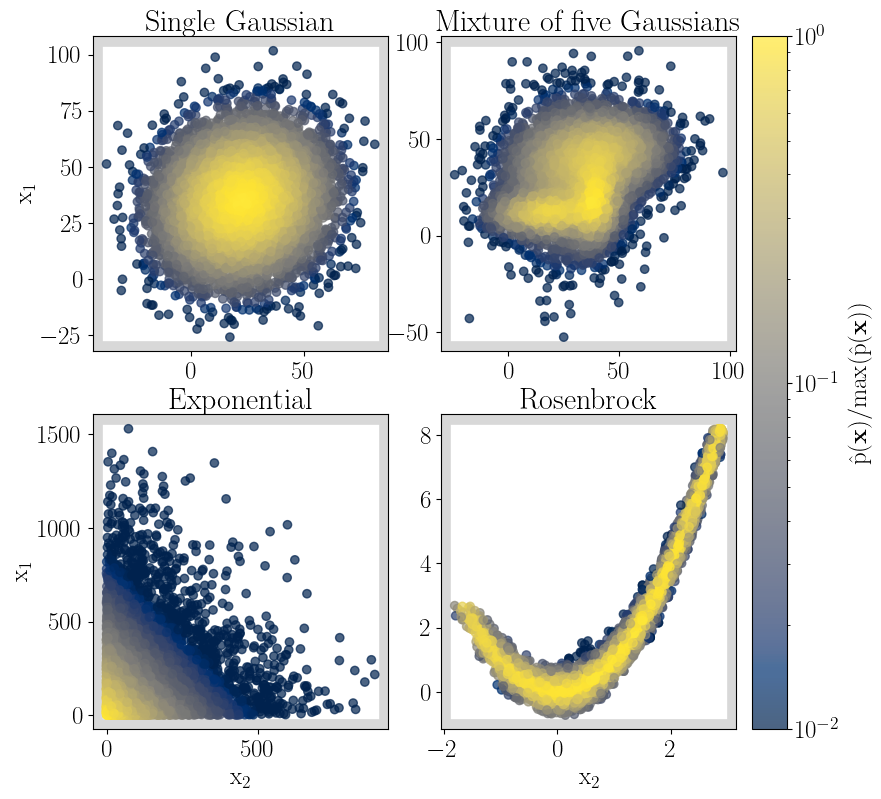}
    \caption{For each unnormalized posterior, we display $10^4$ posterior samples for the case $d=2$. For ease of comparison, the unnormalized posterior $\hat{\rm{p}}(\bf{x})$ is scaled by its maximum and shown in the common color bar. The shaded grey region represents the boundary of the rectangular prior.}
    \label{fig:Comparison_samples}
\end{figure*}

Figure~\ref{fig:Comparison_samples} shows examples of the four distributions in 2-dimensions. The Gaussian distribution has mean $$\bm{\mu} = \begin{pmatrix} 23 \\ 35 \end{pmatrix},$$ and covariance 
$$\bm{\sigma} = \begin{pmatrix} 299 & 31 \\ 31 & 284 \end{pmatrix}.$$
The mixture of five Gaussians has means
$$\bm{\mu}_j = \begin{pmatrix} 39 \\ 19 \end{pmatrix},\begin{pmatrix} 30 \\ 38  \end{pmatrix},\begin{pmatrix} 18 \\ 12 \end{pmatrix},\begin{pmatrix} 46 \\ 44 \end{pmatrix},\begin{pmatrix} 28 \\ 28 \end{pmatrix},$$
and covariance matrices
\begin{align}
\bm{\sigma}_j = &{\begin{pmatrix} 29 & 8 \\ 8 & 118 \end{pmatrix}},{\begin{pmatrix} 250 & 15 \\ 15 & 171 \end{pmatrix}},{\begin{pmatrix} 152 & 4 \\ 4 & 32 \end{pmatrix}},\nonumber\\ &{\begin{pmatrix} 173 & 12 \\ 12 & 107 \end{pmatrix}},{\begin{pmatrix} 198 & 17 \\ 17 & 468 \end{pmatrix}} \,, \nonumber
\end{align}
respectively. The exponential distribution has $$\bm{\lambda} = \begin{pmatrix} 0.009057 \\ 0.005257 \end{pmatrix}.$$ The Rosenbrock case has A = 100, B = 20.

\begin{figure*}
    \centering
    \includegraphics[width=.6\linewidth]{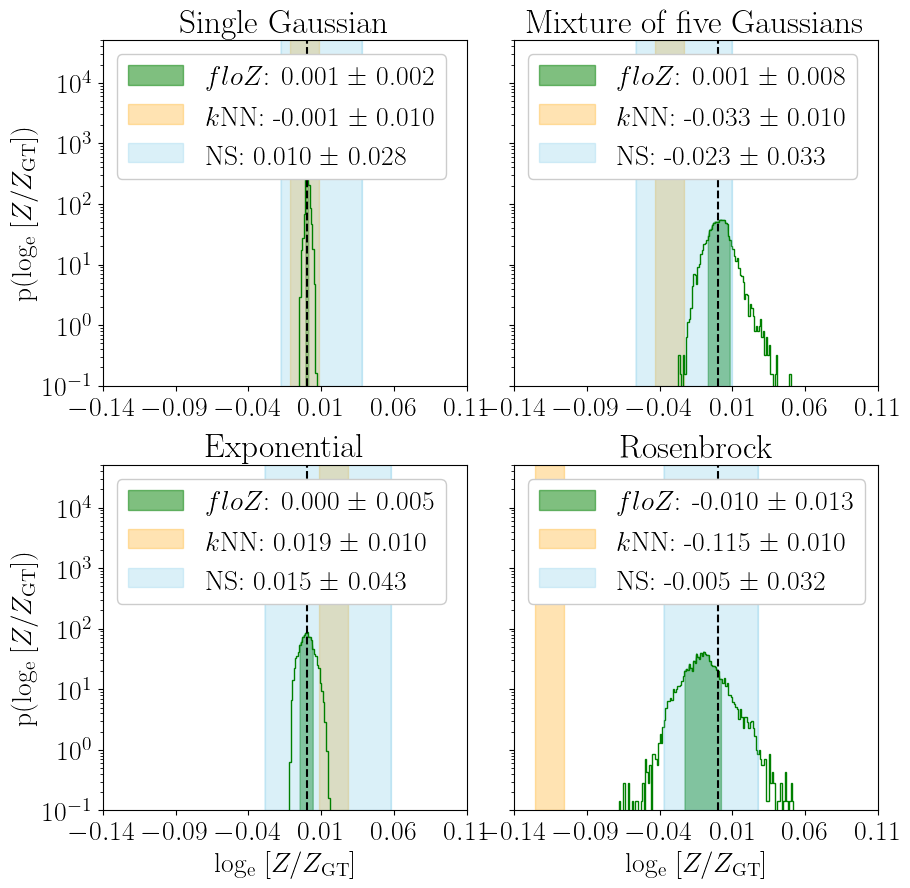}    
    \caption{Evidence estimation in $d=2$ dimensions for (clockwise from top-left): multivariate Gaussian; finite multivariate Gaussian mixture; Rosenbrock; Exponential. In all cases, {\it floZ} and $k$NN employ $10^4$ posterior samples. The true value, represented by the dashed line, has been rescaled to 0, and the shaded regions represent the 1-$\sigma$ uncertainty.}
    \label{fig:Example}
\end{figure*}

\begin{figure*}
    \includegraphics[width=.5\linewidth]{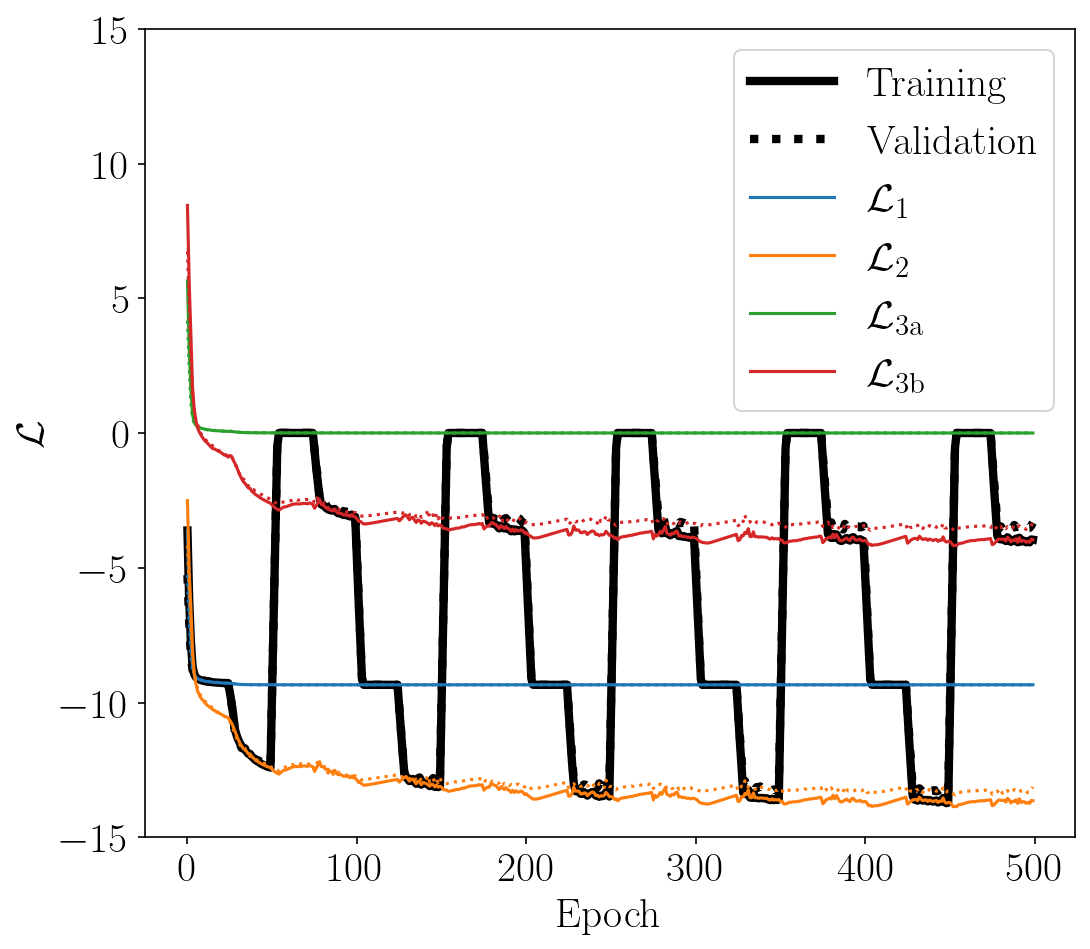}
     \caption{The evolution of the network's loss as a function of training epochs for the case of a 2-dimensional mixture of five Gaussians, illustrates the loss schedule. The four loss terms are shown in color, the total loss in thick/black, and the training (validation) losses are shown in solid (dotted) lines.}
    \label{fig:Loss}
\end{figure*}

\subsection{Benchmarking}
\label{sec:results:benchmark}

\begin{figure*}
    \centering
    \includegraphics[width=.7\linewidth]{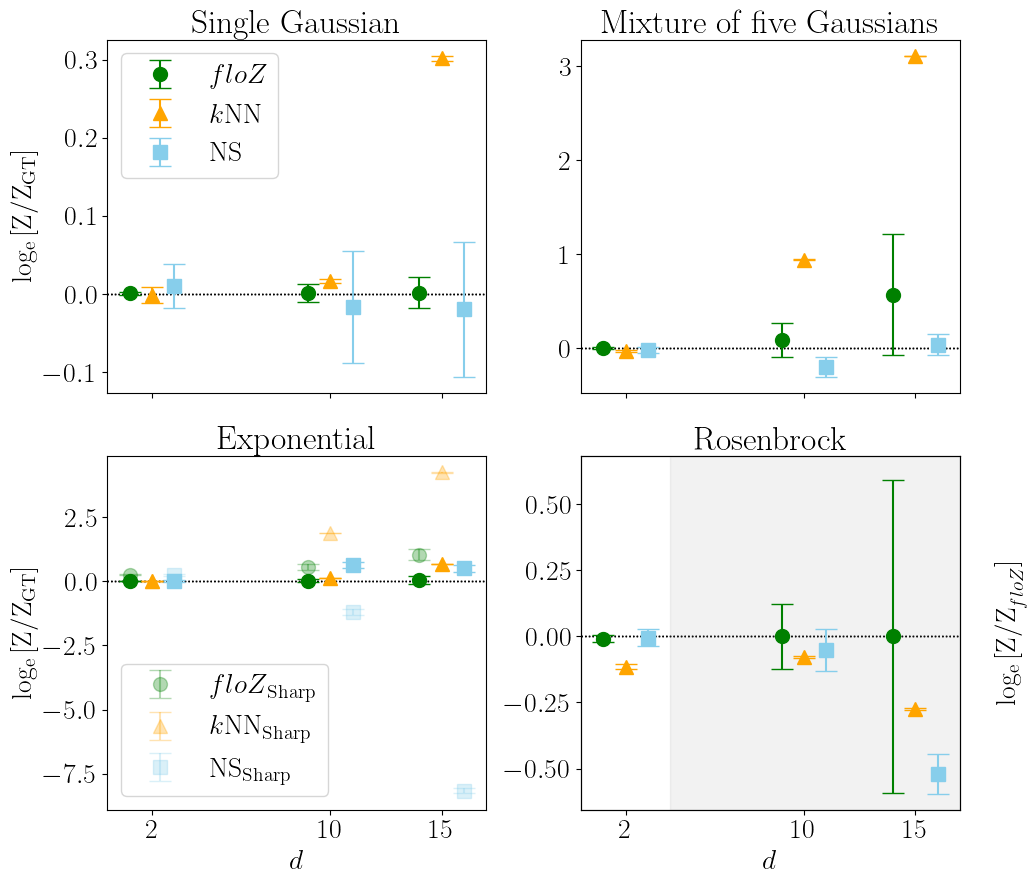}
    \caption{The relative error in the log evidence estimation for different distributions in each panel as a function of the dimensionality of parameter space, $d$. We compare the results of {\it floZ} and $k$NN using the same MCMC samples. We also show the evidence from nested sampling (labeled NS). For the exponential case, the results with solid shading were computed after processing the samples around the sharp edges, whereas those with translucent shading were computed with the sharp distribution edges (labeled 'Sharp'). Since the ground truth for the higher-dimensional \Rosen likelihoods (grey-shaded region) is not numerically tractable, we compare the relative deviation from the {\it floZ} mean prediction (shown by the secondary y-axis).}
    \label{fig:Comparison}
\end{figure*}

\begin{figure}
    \centering
    \includegraphics[width=.9\linewidth]{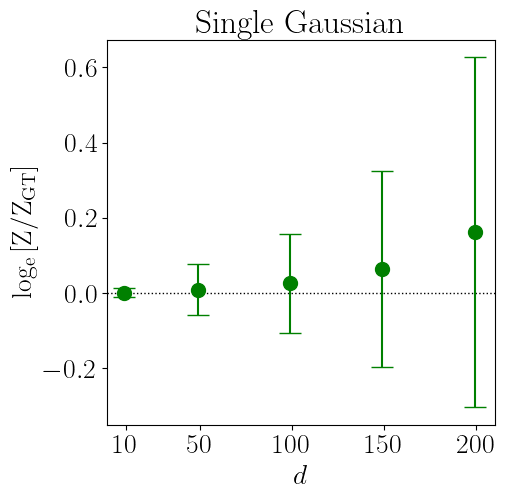}
    \caption{\textit{High dimensional scalability}: the relative error in the log evidence estimation for the single Gaussian case as a function of the dimensionality of parameter space, $d$. For all dimensions, we use $10^5$ posterior samples. We use 11 hidden layers for $d=10$ and 20 hidden layers for higher dimensions.}
    \label{fig:HighDimScaling}
\end{figure}

Figure~\ref{fig:Example} compares the evidence estimate from $floZ$, $kNN$, and nested sampling for the four 2-dimensional posterior distributions. In Figure~\ref{fig:Loss}, we illustrate the behavior of the loss schedule for the Gaussian mixture case. The four individual loss terms are shown with different colors, the total loss in thick/black; training losses are depicted with continuous lines and validation losses with dotted ones. As described in Sec.~\ref{sec:method:algorithm}, we show that each loss term solely contributes to 20 epochs before linearly transitioning to the next term in 5 epochs. In a span of 500 epochs, the network is trained on 5 cycles of the loss terms.

Figure~\ref{fig:Comparison} compares the evidence evaluated by {\it floZ}, $k$NN, and nested sampling (labeled NS) techniques for the four distributions, each simulated in 2, 10, and 15 dimensions. For $d=10$ and $d=15$, we evaluate the evidence using $10^5$ samples with our method and the $k$NN technique, whereas nested sampling uses between 5 to $9\cdot 10^4$ samples. For the exponential case, due to its sharp distribution. we show the results with (solid colors) and without (translucent colors) our prescribed method of sample translation about sharp edges (Sec~\ref{sec:methods:flowDrawbacks}).

The single Gaussian scenario has the simplest distribution and as a result, all methods perform well. However, with increasing complexity and higher dimensions, {\it floZ} and nested sampling are both more accurate than $k$NN.

The exponential distribution highlights the challenge that sharp distributions pose to accurate density and evidence estimation, exacerbated by increasing dimensions. All three techniques show marked improvement upon performing the translation. In addition to a more accurate estimation, the error bars of $floZ$ decrease.

The \Rosen likelihood provides an example with intractable evidence. The ground truth of the 2-dimensional case can be easily computed numerically, however, for higher dimensions, the numerical computation becomes too expensive. We show that for 10 and 15 dimensions, nested sampling and {\it floZ} are in agreement with each other within the estimated error bars.

It is difficult to compare the computational effort required by {\it floZ} versus nested sampling, as the former takes as input posterior samples previously generated. Moreover, the efficiency of nested sampling depends on the details of the algorithm used, as well as on its settings (typically, the number of live samples and tolerance adopted).
For what concerns {\it floZ}, the $d=15$ results of this paper were obtained in less than 10 minutes on an A100 GPU.

In Figure~\ref{fig:HighDimScaling}, we demonstrate the scalability of $floZ$ up to 200 dimensions in the case of a single multivariate Gaussian, whose evidence has been estimated with a constant number of posterior samples, $10^5$ for all cases. The single Gaussian is of course the easiest target for normalizing flows, as the transformation onto the latent space is trivial; more challenging posteriors are not expected to train as well in many dimensions, and would require a larger number of training examples. However, for mildly non-Gaussian distributions of the kind often considered in cosmological data analysis, we expect that {\it floZ} can deliver accurate evidence estimation for parameter space of up to $O(100)$ dimensions. 

\section{Application to real data: gravitational wave analysis}
\label{sec:GW}

\begin{figure*}
    \centering
    \includegraphics[width=.9\linewidth]{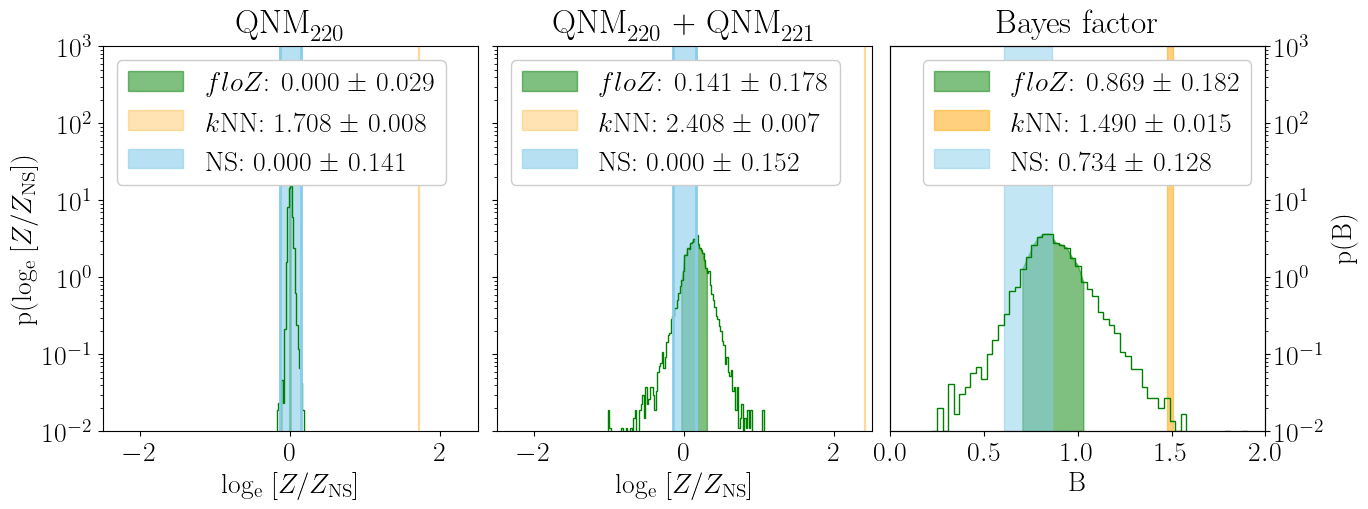}
    \caption{The relative log evidence estimation for the fundamental mode ($\mathrm{QNM}_{220}$) only, left, and also the first overtone ($\mathrm{QNM}_{220}+\mathrm{QNM}_{221}$), center. The evidence is rescaled by the mean of the nested sampling (NS) estimate. The Bayes factor (B) in favor of the presence of the first overtone is shown on the right.
    The shaded regions represent the $1 \sigma$ uncertainty.}
    \label{fig:pyRing}
\end{figure*}

As a practical application of our method in astrophysics, we investigate the presence of the first overtone in the ringdown phase of the first GW detection GW150914 \cite{GW150914}. The ringdown signal from a binary black hole merger can be decomposed as a superposition of quasi-normal modes (QNMs), whose complex frequencies are uniquely determined by the final mass and spin of the remnant black hole. 
Therefore, the observation of more than a single QNM allows for performing consistency tests of General Relativity~\cite{1980ApJ...239..292D,Dreyer:2003bv,Berti:2005ys}.

While it is established that GW150914 contains the fundamental ($n=0$, $l=m=2$) mode \cite{2016PhRvL.116v1101A}, there are conflicting claims regarding the presence of the first overtone ($n=1$, $l=m=2$). Ref.~\cite{2019PhRvL.123k1102I} found the first overtone to be present with a $3.6 \sigma$ confidence. On the contrary, Ref.~\cite{2022PhRvL.129k1102C} found no evidence in favor of an overtone in the data, with a Bayes factor around the waveform peak lower than one. The discussion between the authors of \cite{2019PhRvL.123k1102I} and \cite{2022PhRvL.129k1102C}
is still ongoing \cite{Isi:2022mhy,Isi:2023nif,Carullo:2023gtf}, and the debate is still unsettled. In the meantime, independent analyses (using different techniques) provided different results: Refs.~\cite{2022PhRvD.106d3005F, Crisostomi:2023tle, Wang:2023mst} reported a lower significance for the presence of the overtone (which does not reach $2 \sigma$); Ref.~\cite{Wang:2023xsy} found the Bayes factor for the overtone to lie between 10 and 26; Ref.~\cite{Correia:2023bfn} a Bayes factor of 2.3; and Ref.~\cite{2023PhRvD.107h4010M} a Bayes factor of 600.

These discrepant results can be attributed to the sensitivity of the analysis to different assumptions such as the detector noise characterization, ringdown start time, length of the signal to analyze, marginalizing vs.~fixing some of the parameters, and presence of non-linear effects.
The aim of this section is neither to address the aforementioned issues, nor to provide yet another claim regarding the evidence of the first overtone in the data (or lack thereof), but  we rather aim to test the performance of {\it floZ} in a real scientific case of interest.

We, therefore, use {\it floZ} to compute the evidence for the two models: one where the signal only contains the fundamental mode, and the other where the signal also contains the first overtone.
We obtain posterior samples (and their likelihood) running the ringdown package {\tt pyRing} \cite{pyRing, Carullo:2019flw} for 0.1 seconds of LIGO data at sample rate of 4096 Hz
and with starting time at the LIGO Hanford detector $t_{\text{peak}}=1126259462.42323$ GPS (as estimated in \cite{2022PhRvL.129k1102C}).
We fix the sky position of the source, inclination and polarization as done in \cite{2019PhRvL.123k1102I}, and assume uniform priors on all parameters, with $M_f \in [50, 100] \, M_\odot$, $\chi_f \in [0, 1]$, $A_{22n} \in [0, 5] \cdot 10^{-20}$ and $\phi_{22n} \in [0, 2\pi]$ (where $M_f$ and $\chi_f$ are the final mass and spin, and $A_{22n}$ and $\phi_{22n}$ the amplitude and phase of the QNMs).

{\tt pyRing} relies on the nested sampling algorithm {\tt CPNest}~\cite{cpnest}, which also provides the evidence estimation we use to compare to our for validation (for comparison, we also estimate the evidence with the $k$NN technique).
We used 4096 live points, 4096 maximum Markov Chain steps, and a tolerance of 0.01. For the one-mode model {\tt CPNest} provides 17250 samples, instead for the two-mode model 20000 samples.
Since the posterior distribution for the phase of the fundamental mode peaks close to the boundaries, 
following the method of periodic translations described in Sec~\ref{sec:methods:flowDrawbacks}, we shift this parameter to make the distribution unimodal.

Figure~\ref{fig:pyRing} shows the evidence estimated by {\it floZ} and $k$NN scaled by the result from {\tt CPNest}. The Bayes factor in favor of the presence of the first overtone, as obtained by $floZ$, is $0.87 \pm 0.16$.
For comparison, nested sampling gives $0.76 \pm 0.15$ and $k$NN $1.49 \pm 0.02$. The estimates obtained via nested sampling and {\it floZ} are compatible within their $1 \sigma$ uncertainties.

As a final note, let us stress that although in this particular case the use of {\it floZ} would seem unnecessary since a nested sampler is already available, the running time of {\tt pyRing}, on 4 Intel(R) Xeon(R) Gold 6238R CPU @ 2.20GHz, was of the order of several hours, which should be compared to the few minutes required by {\it floZ}. Therefore, for ringdown analyses, our tool could be better paired with the latest implementation of {\tt ringdown} \cite{maximiliano_isi_2021_5094068, Isi:2021iql} on a {\tt JAX} back-end (which supports GPU execution) and the Hamiltonian Monte Carlo (HMC) sampler in its {\tt NumPyro} implementation, which could supply posterior samples from the unnormalized target for fast evidende estimation with {\it floZ}.

\section{Conclusions} 
\label{sec:conclusions}
We have introduced {\it floZ}, a novel method that uses normalizing flows for estimating the Bayesian evidence and its numerical uncertainty using pre-existing samples drawn from the target unnormalized posterior distribution, e.g. as obtained by standard MCMC methods.

Normalizing flows are built to map a complex target probability function (in our case, the unnormalized posterior distribution, given by the product of a likelihood and prior)
into a simple probability distribution (e.g. a Gaussian). The transformation between the original variables and the latent ones is bijective and can be modeled with a neural network. 
We exploit the fact that the flows encode the volume of the real-space distribution to compute the evidence from a set of posterior samples previously gathered by any suitable algorithm (e.g., MCMC), while at the same time reducing the estimate's scatter thanks to three novel additional loss terms for the training of the normalizing flow. Furthermore, we address the limitation of modeling distribution with sharp boundaries with a pre-processing technique showing marked improvement in the results from normalizing flows and other state-of-the-art evidence estimation techniques that use nested sampling and $k$-nearest-neighbours.

We have validated {\it floZ} against analytical benchmarks in parameter spaces of up to 15 dimensions: multivariate Gaussians, a finite multivariate Gaussian mixture, multivariate exponential distribution, as well as, the \Rosen distribution. Our method demonstrates performance comparable to nested sampling (which requires to be run ad hoc) and superior to a $k$-nearest neighbors method employed on the same posterior samples. We believe that {\it floZ} will add to the toolbox of astronomers and cosmologists seeking a fast, reliable method to compute the evidence from existing sets of posterior samples. {\it floZ} will be especially useful for cases where the likelihood is expensive and a full nested sampling run is difficult to achieve, while MCMC samples can be obtained more efficiently thanks to the ease of parallelization of many MCMC algorithms (such as Gibbs sampling). 

Pushing our approach to much larger parameter spaces remains challenging because of two limitations, both fundamentally stemming from the curse of dimensionality: first, obtaining posterior samples in high dimensions becomes more difficult (although some methods, like Gibbs sampling and HMC, can show mild scaling with dimensionality in some favorable circumstances); second, training the flow in higher dimensions suffers from increasing inaccuracies (but see \cite{Piras2024TheFO} for a normalizing-flows-enhanced harmonic mean estimator with similar performance as our own up to 150 dimensions, at least for the restricted case of unimodal, mildly non-Gaussian targets). However, we note that we only need accurate densities in latent space near the peak (as opposed to into the tails of the distribution), which opens the door to potentially exploiting the existence of a well-defined `typical set' in high dimensions, a consequence of the concentration of measure phenomenon. 
An interesting possibility for improvement over the aforementioned inaccuracies of the trained flow, potentially allowing for enhanced precision, is provided by {\it importance sampling} \cite{2023PhRvL.130q1403D}, where importance weights are applied to correct the distribution of the estimated evidence, based on the underlying posterior distribution and its recovery with the flow.
We shall explore these ideas further in a dedicated future paper. 

To demonstrate the use of $floZ$ on real data, we estimated the Bayes factor of the presence/absence of the first overtone in the ringdown phase of the first GW observation by the LIGO-Virgo collaboration \cite{GW150914}, namely GW150914. Our estimate is consistent with that of nested sampling. Among other applications of  {\it floZ} to gravitational-wave astronomy, it is worth mentioning the recent evidence for a stochastic signal reported by pulsar timing array experiments~\cite{Antoniadis:2023ott,Tarafdar:2022toa,NANOGrav:2023gor,Reardon:2023gzh,Xu:2023wog}. Within a Bayesian framework, interpreting the nature of this stochastic signal requires comparing different hypotheses and their evidences, which {\it floZ} could readily compute from the existing samples released by the experiments. Similarly, as another of the many possible applications, {\it floZ} could be used to investigate the statistical robustness of features/peaks that may be present in the mass function of the astrophysical black holes detected by the LIGO-Virgo-KAGRA collaboration~\cite{KAGRA:2021duu}.

{\bf Data availability:} The {\it floZ} code to reproduce the results in this paper, and to estimate the evidence for any set of posterior samples, likelihood, and prior densities is available at \url{https://github.com/Rahul-Srinivasan/floZ}.

\begin{acknowledgments}
We thank Alan Heavens for insightful discussions and Uro\v{s} Seljak for comments on an earlier draft. 
E.B., R.S. and M.B. acknowledge support from the European Union’s H2020 ERC Consolidator Grant ``GRavity from Astrophysical to Microscopic Scales'' (Grant No. GRAMS-815673), the PRIN 2022 grant ``GUVIRP - Gravity tests in the UltraViolet and InfraRed with Pulsar timing'', and the EU Horizon 2020 Research and Innovation Programme under the Marie Sklodowska-Curie Grant Agreement No.~101007855. 
M.C. is funded by the European Union under the Horizon Europe's Marie Sklodowska-Curie project~101065440.
R.T. acknowledges co-funding from Next Generation EU, in the context of the National Recovery and Resilience Plan, Investment PE1 – Project FAIR ``Future Artificial Intelligence Research''. This resource was co-financed by the Next Generation EU [DM 1555 del 11.10.22]. RT is partially supported by the Fondazione ICSC, Spoke 3 ``Astrophysics and Cosmos Observations'', Piano Nazionale di Ripresa e Resilienza Project ID CN00000013 ``Italian Research Center on High-Performance Computing, Big Data and Quantum Computing'' funded by MUR Missione 4 Componente 2 Investimento 1.4: Potenziamento strutture di ricerca e creazione di ``campioni nazionali di R\&S (M4C2-19 )'' - Next Generation EU (NGEU).
\end{acknowledgments}

\bibliographystyle{utphys}
\bibliography{refs}

\end{document}